\documentclass[11pt,letter]{article}

\usepackage[letterpaper,top=.13in,bottom=.17in,left=.57in,right=.57in,includefoot]{geometry}
\usepackage[natbib=true,style=nature,sorting=none]{biblatex} 
\usepackage{graphicx}
\usepackage{booktabs}
\usepackage{fancyhdr}
\usepackage[fleqn]{amsmath}
\usepackage{epstopdf}
\usepackage{amsfonts}
\usepackage{setspace}
\usepackage{verbatim}
\usepackage{rotating}
\usepackage{multirow}
\usepackage{txfonts}
\usepackage{blindtext}
\usepackage{multicol}
\usepackage{caption}
\AtBeginEnvironment{tabular}{\small}
\usepackage{enumitem}
\setlist[itemize]{leftmargin=*,topsep=0pt,itemsep=-1ex,partopsep=1ex,parsep=1ex}

\usepackage[compact]{titlesec}
\titleformat*{\section}{\MakeUppercase}

\titleformat*{\subsection}{\it}
\titleformat*{\subsubsection}{\it}

\begin{document}\pagestyle{empty}
\parbox[][1.3in][t]{\textwidth}{%
  {\LARGE Continual Person Identification using Footstep-Induced Floor Vibrations on Heterogeneous Floor Structures\\~\\} 
  {\Large Yiwen Dong \& Hae Young Noh}\\
  \emph{\large Stanford University, California, U.S.A.}\medskip
  
}

\vspace{-0.1in}
\parbox[][2.5in][t]{\textwidth}{%
ABSTRACT: Person identification is important for smart buildings to provide personalized services such as health monitoring, activity tracking, and personnel/customer management. However, previous person identification relies on pre-collected data from everyone, which is impractical in many buildings and public facilities in which visitors are typically expected. This calls for an online person identification system that gradually learns people’s identities on the fly. Existing studies use cameras to achieve this goal, but they require direct line-of-sight and also have raised privacy concerns in public. Other modalities such as wearables and pressure mats are limited by the requirement of device-carrying or dense deployment. Thus, prior studies introduced footstep-induced structural vibration sensing, which is non-intrusive and perceived as more privacy-friendly. However, this approach has a significant challenge - the high variability of vibration data due to structural heterogeneity and human gait variations, which makes online person identification algorithms perform poorly. In this paper, we characterize the variability in footstep-induced structural vibration data for accurate online person identification. To achieve this, we quantify and decompose different sources of variability and then design a feature transformation function to reduce the variability within each person’s data to make different people's data more separable. We evaluate our approach through field experiments with 20 people. The results show a 70\% variability reduction and a 90\% accuracy for online person identification.
}

\begin{multicols*}{2}

\section{Introduction}

Structural vibrations induced by humans contain various information, including their identities, activities, and health status [4, 5, 7]. Among them, identity is essential for smart buildings as it is the premise of personalized services, including health monitoring, personnel management, and emergency assistance for patients and the elderly. Person identification typically relies on pre-collected data from the occupants. However, in many real-life scenarios, it is impractical to collect everyone’s data, especially when visitors are frequently present [3]. This calls for an online person identification system that gradually learns people’s identities as it observes more data over time. 

However, existing online person identification systems have limitations in both sensing modalities and learning methods. Previous studies use cameras for online person identification, but they are not suitable for complex indoor spaces due to the direct line-of-sight requirement [9]. It also has raised privacy concerns due to appearance exposure. Other sensing modalities, such as wearables and pressure mats [6, 11], could be used to reduce these concerns, but they have limited scalability due to device-carrying and dense deployment requirements. Thus, previous studies introduced a person identification system based on footstep-induced structural vibration sensing [7], which needs only sparsely deployed sensors, is non-intrusive and is perceived as more privacy-friendly. Since different people walk differently, their footstep-induced floor vibrations captured by the sensors are also unique, which enables person identification. However, one significant challenge of this approach is the high variability of vibration data due to structural heterogeneity and human gait variations, which makes the existing online person identification algorithms perform poorly.

In this paper, we characterize the variability in footstep-induced structural vibration data to develop a feature transformation approach that enables online person identification. We transform the footstep data into a new feature space where the within-person variability is reduced while the between-person separability is enhanced. To achieve this, we decompose the sources of variability, quantify the variability based on the covariance between footsteps, and design the transformation function based on the dominant variability source. We then formulate an optimization problem that aims to find the transformation parameters which map features to the new space with minimal variability and maximum separability. With these transformed features, we develop a non-parametric Bayesian online learning approach based on Dirichlet Process that detects and learns the newcomers’ footstep features and then updates the overall footstep feature model on the fly. We evaluate our approach through field experiments with 20 people across 2 structures. For both structures, our method reduces the feature variability by 70\% compared to the original data. Our method also achieves a 90\% average accuracy in identifying 10 people on each structure in an online manner starting from 1 person's data.

\section{Variability Analysis in Footstep-Induced Structural Vibrations}
In this section, we analyze the variability in footstep-induced structural vibration data in order to understand and model the variability. The footstep-induced structural vibrations are highly variable due to the natural variations in human gait (footstep variability) and the structural heterogeneity (structural variability). Therefore, we decompose and quantify the sources of variability to design a variability reduction method.

\subsection{Variability Decomposition}
We decompose the variability in footstep-induced structural vibrations into two main sources: 1) footstep variability and 2) structural variability. Figure 1 shows the significant difference between the same person’s footstep signals due to these two sources. The footstep variability results from the natural variations in human gait. These variations may come from the minor adjustments of balance when a person is walking in the same direction at a relatively constant speed [2]. In addition, the structural heterogeneity also contributes to the overall variability in footstep-induced floor vibrations [10]. The structural variability is defined as the signal variations caused during the wave generation and the wave propagation process from the excitation locations to the sensors.

\includegraphics[width=3.53in]{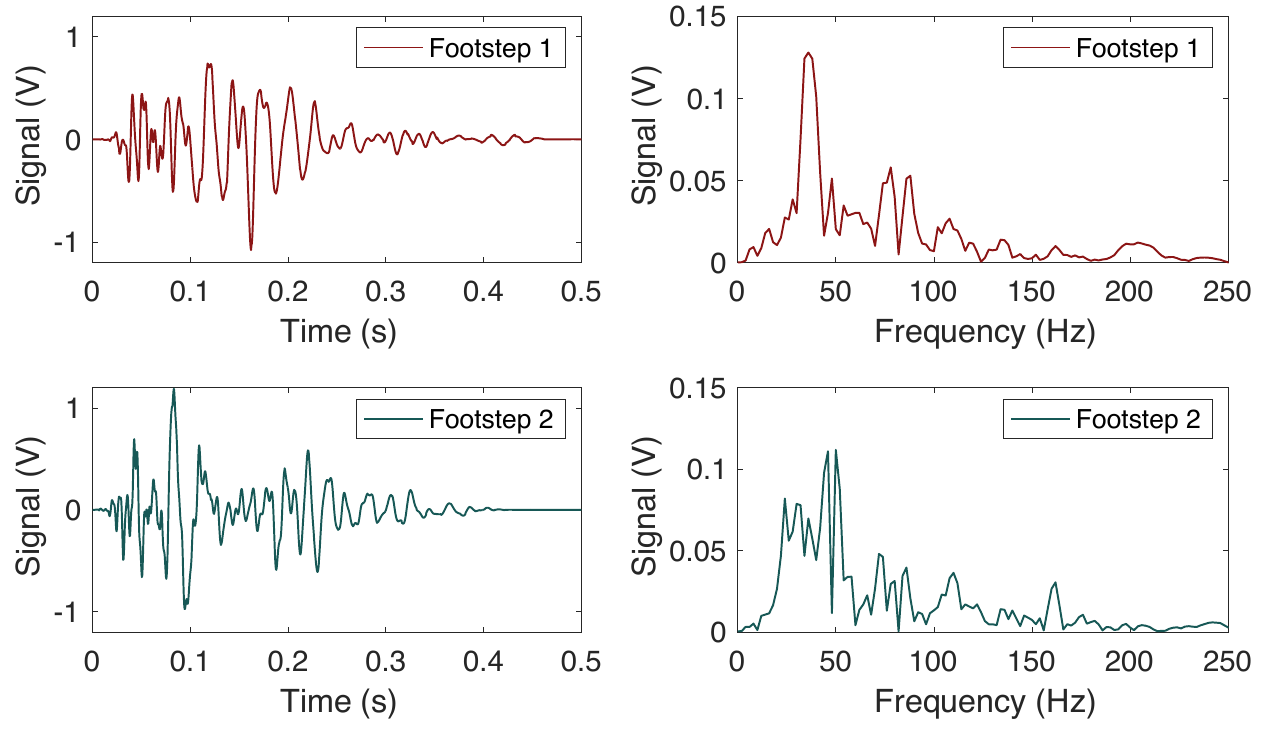}
\captionof{figure}{The same person induces different structural vibrations, visualized in time- and frequency- domain.}\label{fig:1}

To decompose the structural and footstep variability, we compare the vibration responses from footstep impact forces and the ball drop forces from the same height at the same excitation and sensing location. To isolate the footstep variability, we use ball drops at the same height, which gives almost identical input forces. To remove the structural variability, we set the same group of excitation and sensing locations, which allows the wave generation and propagation paths to be almost the same.

\includegraphics[width=3in]{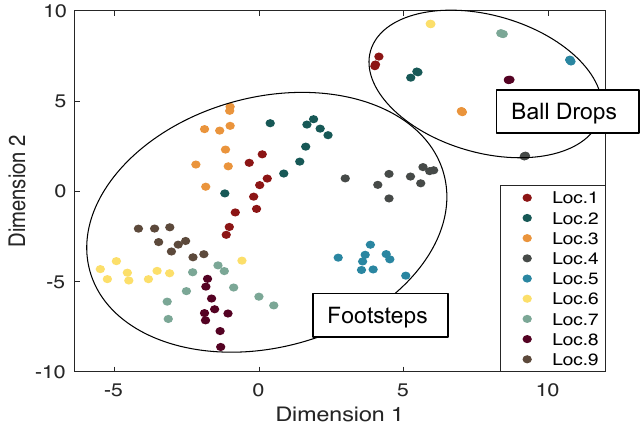}
\captionof{figure}{Variability decomposition using 2 types of impulses: 1) ball drops, 2) footsteps. Different colors indicate different excitation locations, numbered as 1-9.}\label{fig:2}

In Figure~\ref{fig:2}, we observe that the structural variability has a much larger variability than the footstep variability because the average Euclidean distance between different excitation locations in the feature space is larger than that of different footsteps at the same excitation location.

\subsection{Variability Quantification}
To understand how much structure heterogeneity and footstep variations contribute to the overall variability, we quantify these decomposed variability sources through variability analysis of the frequency-domain footstep features. Since the footstep impact forces are generated from repeated trials of the same person with consistent body mechanism, they are assumed to be independent and identically distributed (i.i.d.). Therefore, we model the footstep variability as a statistically random factor incurred during sampling (i.e., walking in our problem), which is defined as follows:
\begin{equation}
    \Sigma_{footstep} = \frac{1}{K} \sum_{k=1}^{K} \frac{1}{N_k} \sum_{i=1}^{N_k} (x_i - \mu_k)(x_i - \mu_k)^T
    \label{eq:footstepvar}
\end{equation}

The structural variability, on the other hand, is estimated by sampling uniformly distributed excitation locations over the structure and then computing the covariance among the sample means at each location. As observed in Figure~\ref{fig:2}, the mean of footstep features at each location are scattered around the feature space and can be modeled as an overall normal distribution. They also satisfy the i.i.d. assumption because they are generated through the same structure that is typically stationary without damage. Therefore, it is quantified as follows:
\begin{equation}
    \Sigma_{structure} = \frac{1}{K} \sum_{k=1}^{K} (\mu_k - \mu)(\mu_k - \mu)^T
    \label{eq:structurevar}
\end{equation}
where $K$ denotes the total number of excitation locations, $N_k$ denotes the number of footsteps at location $k$. $x_i$ means the features of $i$-th footstep sample at location $k$, $\mu$ is the mean among all footstep features and $\mu_k$ is the mean of footstep features at location $k$.

Based on the above quantification of structural and footstep variability, we compare the variability proportions on the wood and concrete structures and found that the structural variability dominates (See Figure~\ref{fig:3}). 
\includegraphics[width=3.1in]{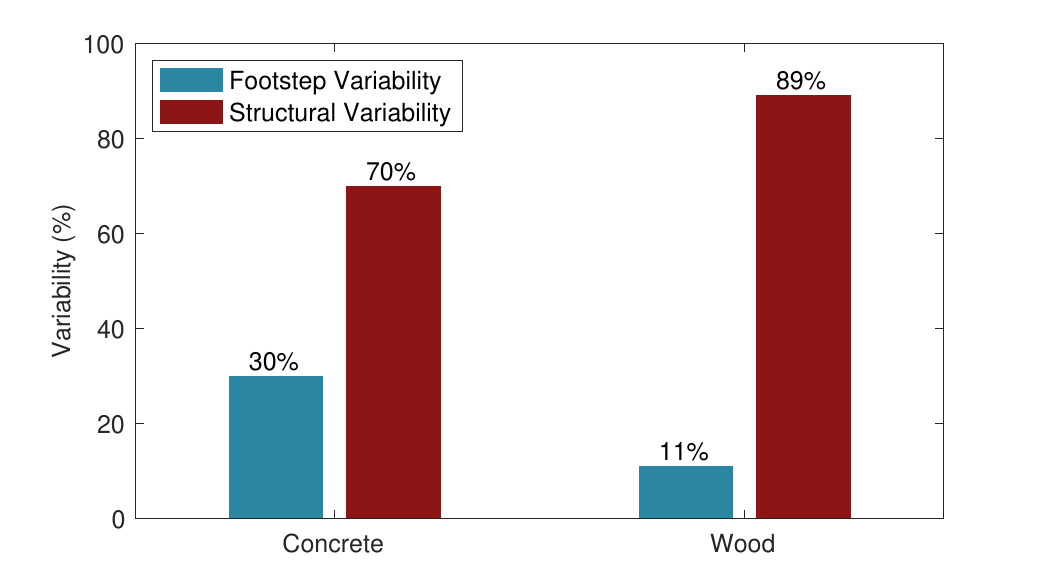}
\captionof{figure}{Overall variability proportion for wood and concrete structures.}\label{fig:3}

\section{Variability Reduction Via Optimization-based Feature Transformation}
In order to reduce the dominant structural variability, we design an optimization-based feature transformation function to develop more separable footstep features between people.

\subsection{Physics-Guided Transformation Function Design}
Since the structural variability is reflected through vibration signals with different excitation and sensing locations, we explore the relationship between these footstep signals and introduce a feature transformation function to describe how different footsteps are mapped across different excitation and sensing locations.

\textbf{Structural Variability due to Wave Generation:} Assuming the cross-sections of our structure can be simplified as a simply-supported beam within the linear elastic range, the dynamic response of the structure can also be simplified, in which we can compute the analytical solution of the structural vibration given the input force and location [8]. Let $P(t)$ be the input force with respect to time, $w(x,t)$ be the vertical deflection of the beam at location $x$, the person's walking speed as $v$, the governing equation for vertical vibration can be written as:
 \begin{equation}
    EI\frac{\partial^4 w(x,t)}{\partial x^4} + \rho A \eta \frac{\partial w(x,t)}{\partial t} + \rho A \frac{\partial^2 w(x,t)}{\partial t^2} = P(t)\delta(x-vt)
    \label{eq:dynamics}
 \end{equation}
 where $E$, $\rho$, $A$, $I$ are material and geometry constants, $\eta$ is the damping coefficient, $\delta(t)$ is the Dirac delta function.

After solving the above equation, the structural responses at 2 different excitation location $x_1$ and $x_0$ can be approximated as follows:
\begin{equation}
    \begin{bmatrix}
    \phi_k(x_1, \omega_1)\\
    \vdots\\
    \phi_k(x_1, \omega_d)
    \end{bmatrix}
    = C
    \begin{bmatrix}
    e^{\lambda(\omega_1)(x_1-x_0)} & \cdots & 0\\
    0 & \ddots & 0\\
    0 & \cdots & e^{\lambda(\omega_d)(x_1-x_0)}
    \end{bmatrix}
    \begin{bmatrix}
    \phi_k(x_0, \omega_1)\\
    \vdots\\
    \phi_k(x_0, \omega_d)
    \end{bmatrix}
\end{equation}
where $\omega_i, i = 1, 2, ..., d$ means different frequency bands, and $C$ is a constant value.

Therefore, we can apply a linear transformation matrix to approximate the relationship of structural responses across different excitation locations.

\textbf{Structural Variability due to Wave Propagation:}
After the vibration waves are generated, they propagate through the structural medium and gradually attenuate as the wave propagation distance gets longer. The amount of attenuation in the vibration waves depends on 1) the traveling distance $l$, 2) the attenuation coefficient $\alpha$, and 3) the frequency $\omega$ at which it propagates [1]. The function can be written as:
 \begin{equation}
 \frac{I_1}{I_0} = e^{\alpha l \omega}
 \end{equation}
 where $I_0$ and $I_1$ represents the signal intensity (i.e., energy) at the beginning and the end of propagation. Since our footstep features (i.e., the frequency-domain signal amplitudes) are the square root of the signal intensity (i.e., $\phi(x, \omega) = \sqrt{I}$), the footstep features at 2 different locations due to wave attenuation can be associated as:
 \begin{equation}
         \begin{bmatrix}
    \phi(x_1, \omega_1)\\
    \vdots\\
    \phi(x_1, \omega_d)
    \end{bmatrix}
    = C
    \begin{bmatrix}
    e^{-\frac{1}{2}\alpha l \omega_1} & \cdots & 0\\
    0 & \ddots & 0\\
    0 & \cdots & e^{-\frac{1}{2}\alpha l \omega_d}
    \end{bmatrix}
    \begin{bmatrix}
    \phi(x_0, \omega_1)\\
    \vdots\\
    \phi(x_0, \omega_d)
    \end{bmatrix}
 \end{equation}
 
From the above equation, the wave attenuation effect in structural vibration from the excitation location and the sensing location is equivalent to a linear transformation. Therefore, we design a linear transformation function to reduce the structural variability, represented as follows:
\begin{equation}
{X}_{transformed} = {w}^{T}{X}_{original}   \end{equation}
Where $X_{original}$ represents the extracted footstep features before the transformation, and $X_{transformed}$ denotes the transformed features after the optimization.

\subsection{Formulation of the Optimization Problem}
 With the designed transformation function, we formulate an optimization problem with an explicit goal to separate different people. The optimization goal is consistent with the new person identification objective, which is to distinguish different people through their footsteps. 
 
 To translate this goal into mathematical expressions, he optimization problem is formulated to minimize the sum of within-person covariance and maximize the between-people covariance, expressed as follows:
\begin{equation}
maximize \quad J(w) = \frac{w^{T}S_{B}w}{ w^{T}S_{W}w }
 \end{equation}
Where the between-person covariance is $S_B = \sum_{i=1}^{C} N_i (\mu_i - m)(\mu_i - m)^T$, and the within-person covariance matrix is ${S}_{W} = \sum_{i=1}^{C} \sum_{n=1}^{N_i} (x_n^i - \mu_i)(x_n^i - \mu_i)^T$. $i$ represents the person number that ranges from 1 to $C$; $n$ represents the sample number within each person $i$ that ranges from 1 to $N_i$ ; $x$ denotes the data sample; $m$ and $\mu_i$ are the mean of all data and the mean of data in person $i$ respectively. Since the optimization goal matches the objective of Fisher’s Discriminant Analysis parameterized by coefficents $w$, the optimal coefficients $w$ are estimated through a closed form solution in the Fisher's formulation [12]. 

\section{Online Person Identification using Dirichlet Process}
After reducing the variability in footstep-induced structural vibrations, we describe the feature distributions among different people as a Dirichlet Process Mixture Model (DPMM) to predict the identity of each observed person. DPMM is a non-parametric Bayesian model that allows us to model and update an unlimited number of visitors. As each person is observed, we first estimate the prior probability of his/her identity based on the previously observed data. We then update the posterior in the model with the footstep data from this newly observed person [13]. Finally, the person is identified as the one with the largest posterior probability. As more people are observed, we update the DPMM with each newly observed data and its prediction. 

\vspace{-0.1in}
\section{Evaluation}
To evaluate the effectiveness of our variability reduction method in online person identification, we conducted field experiments on a wood and a concrete structure with 20 participants. All experiments were conducted in accordance with the approved IRBs.

\subsection{Experiment Setup}
We conducted experiments on 2 deployment sites: 1) a wood-framed platform and 2) a concrete corridor. For the setup, 4 SM-24 geophone sensors recording at a sampling frequency of 25.6 kHz were mounted along the two edges of the corridor, spaced apart by 2 meters (see Figure~\ref{fig:4}). Amplifiers were used to improve the signal-to-noise ratio (SNR) [10]. A National Instruments DAQ was used to acquire and convert the analog signal to the digital signal [7]. The participants were asked to walk across the platform more than 10 times using their natural gait; for each time of walking, a series of consecutive footsteps were recorded.

\includegraphics[width=3.53in]{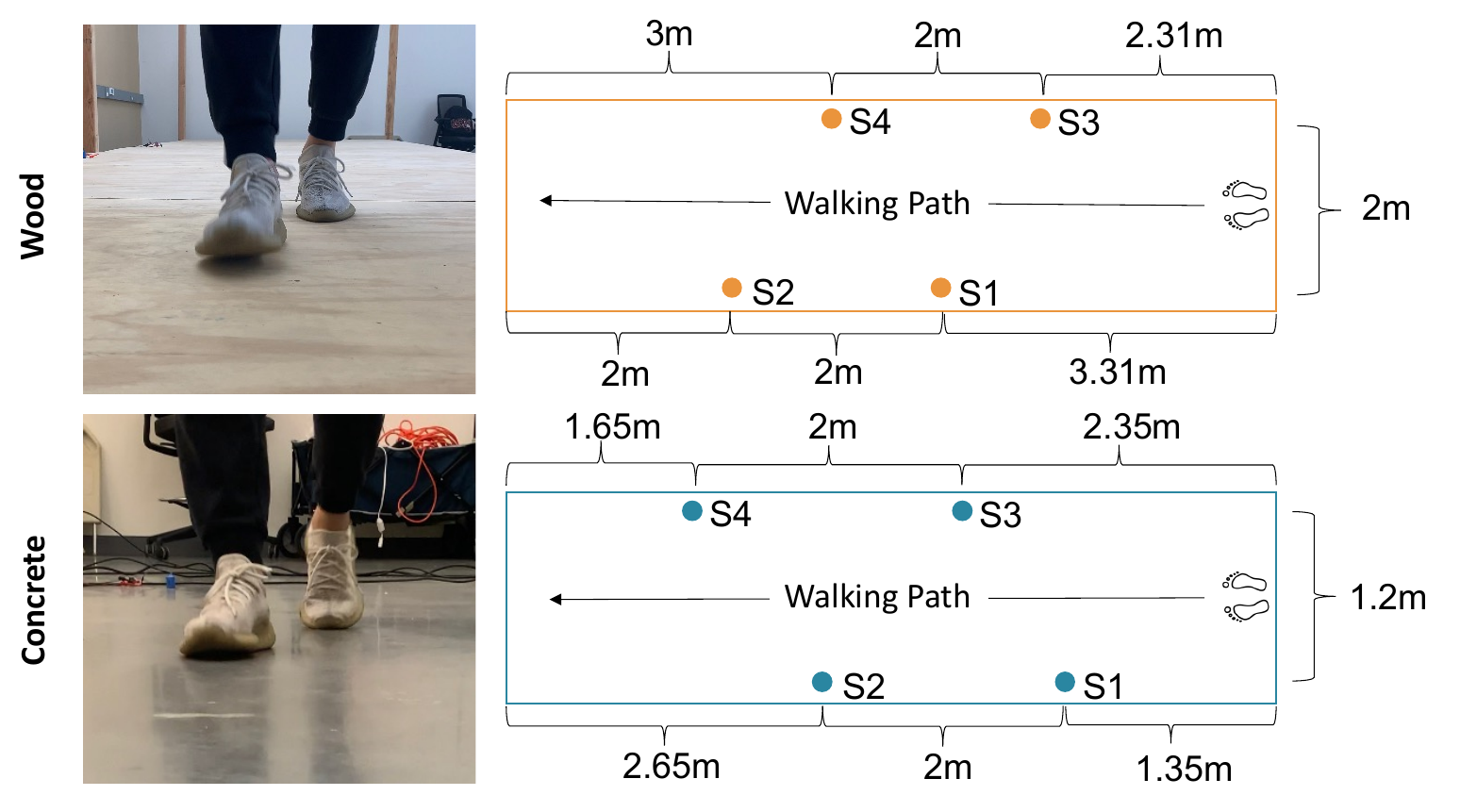}
\captionof{figure}{Experiment layout.}\label{fig:4}

\subsection{Evaluation Results}
Overall, our method achieves an average of 90\% accuracy in online person identification with 1 pre-recorded person only. For both structures, our method reduces the feature variability by 70\% compared to the original data. 

As shown in Figure~\ref{fig:5}, the footsteps between different people (in different colors) are more separable after the feature transformation.

 \includegraphics[width=3.53in]{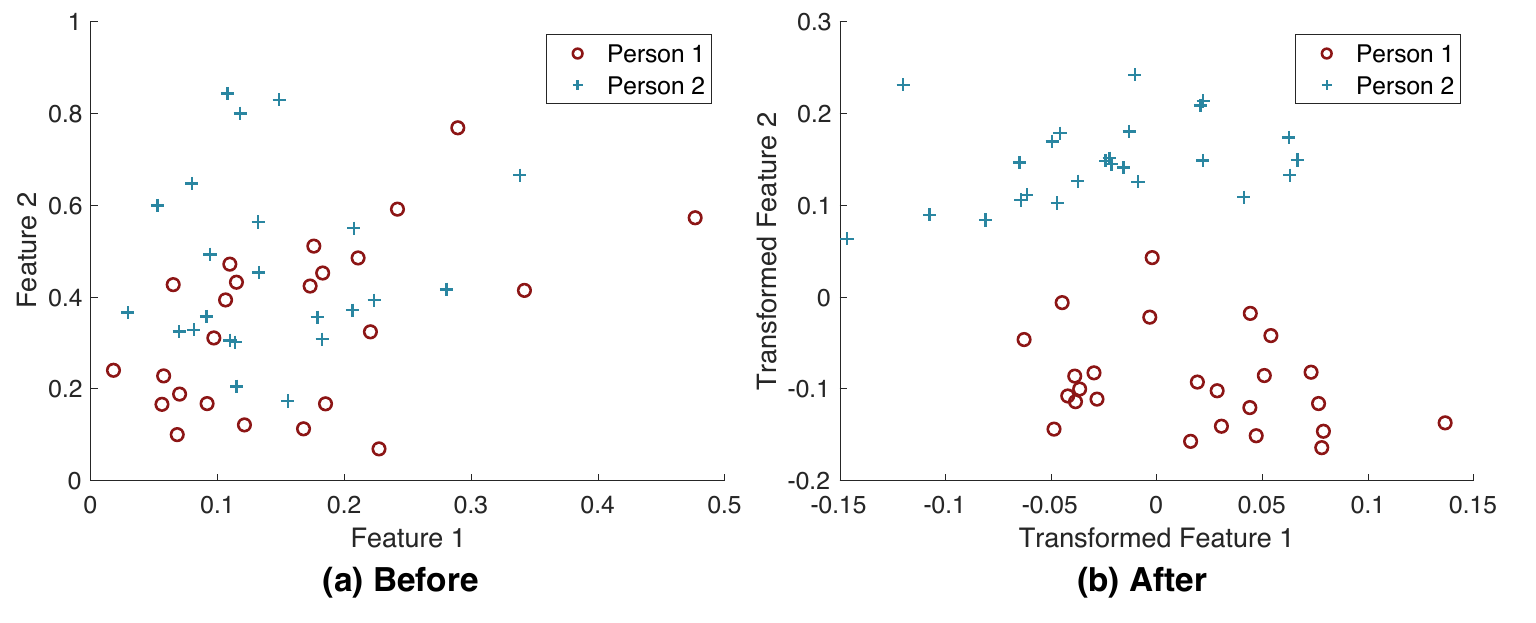}
 \caption{2D t-SNE visualization of footstep features (a) before and (b) after variability reduction for 2 people.}\label{fig:5}

\section{Conclusion}
In this paper, we characterize the variability in footstep-induced structural vibration data to develop a data transformation approach that enables online person identification. Through physics-guided variability analysis, the footstep data are transformed into a new feature space where the within-person variability is reduced while the between-person separability is enhanced. The effectiveness of this transformation is evaluated through real-world walking experiments across 2 different structures. Our method reduces the feature variability by 70\% compared to the original data for both structures, and achieves a 90\% average accuracy in identifying 10 people on each structure starting from 1 person’s data only.

\section{References}
{\fontsize{9pt}{11pt}\selectfont
[1] Ammon, C. J., Velasco, A. A., Lay, T. \& Wallace, T. C. 2021. An overview of earthquake and seismic-wave mechanics. Foundations of Modern Global Seismology, 39–63. \par
[2] Fagert, J. et al. 2021. Structure- and Sampling-Adaptive Gait Balance Symmetry Estimation Using Footstep-Induced Structural Floor Vibrations. Journal of Engineering Mechanics 147(2), 04020151. \par
[3] Dong, Y. et al. 2021. Non-Parametric Bayesian Learning for Newcomer Detection Using Footstep-Induced Floor Vibration: Poster Abstract in Proceedings of the 20th International Conference on Information Processing in Sensor Networks. 404-405. \par
[4] Dong, Y. et al. 2020. MD-Vibe: Physics-informed analysis of patient-induced structural vibration data for monitoring gait health in individuals with muscular dystrophy. UbiComp/ISWC 2020 Adjunct - Proceedings of the 2020 ACM International Joint Conference on Pervasive and Ubiquitous Computing and Proceedings of the 2020 ACM International Symposium on Wearable Computers, 525–531. \par
[5] Dong, Y., et al. 2022. Re-vibe: Vibration-based indoor person re-identification through cross-structure optimal transport. In Proceedings of the 9th ACM International Conference on Systems for Energy-Efficient Buildings, Cities, and Transportation (pp. 348-352).\par
[6] Brutti, A. \& Cavallaro, A. 2017. Online Cross-Modal Adaptation for Audio-Visual Person Identification with Wearable Cameras. IEEE Transactions on Human-Machine Systems 47, 40–51.\par
[7] Pan, S. et al. 2017. FootprintID: Indoor Pedestrian Identification through Ambient Structural Vibration Sensing. Proceedings of the ACM on Interactive, Mobile, Wearable and Ubiquitous Technologies 1, 1–31. \par
[8] Khiem, N. T. \& Hang, P. T. 2016. Frequency response of a beam-like structure to moving harmonic forces. Vietnam Journal of Mechanics 38, 223–238. \par 
[9] Lu, Y. et al. 2015. Online person identification and new person discovery using appearance features. 2015 IEEE International Conference on Evolving and Adaptive Intelligent Systems, EAIS.\par  
[10] Pan, S. et al. 2015. Indoor person identification through footstep induced structural vibration. HotMobile - 16th International Workshop on Mobile Computing Systems and Applications, 81–86. \par 
[11] Suutala J. \& R\"{o}ning J. 2008. Methods for person identification on a pressure-sensitive floor: Experiments with multiple classifiers and reject option. Information Fusion 9, 21–40.\par 
[12] Scholkopft, B. \& Mullert, K. 1999. Fisher Discriminant Analysis with Kernels, 41–48. \par 
[13] Izenman, A. J. 1991. Review papers: Recent developments in nonparametric density estimation. Journal of the American Statistical Association 86, 205–224. \par
}

\end{multicols*}

\end{document}